%% file: neurips_2025.tex
\documentclass{article}

\usepackage[preprint]{neurips_2025}

\usepackage[utf8]{inputenc} 
\usepackage[T1]{fontenc}    
\usepackage{hyperref}       
\usepackage{url}            
\usepackage{booktabs}       
\usepackage{amsfonts}       
\usepackage{nicefrac}       
\usepackage{microtype}      
\usepackage{xcolor}         
\usepackage{graphicx}         
\usepackage[breakable]{tcolorbox}
\tcbuselibrary{listings, skins, breakable}

\usepackage{subcaption}
\usepackage{amsmath}
\usepackage{caption}
\usepackage{amsfonts}
\usepackage{setspace}
\usepackage{url}
\usepackage{multirow}
\usepackage{makecell}
\usepackage{color}
\usepackage{xspace}
\usepackage{adjustbox}
\usepackage{soul}
\usepackage{float}
\usepackage{textcomp}
\usepackage{enumitem}
\usepackage{algorithm}
\usepackage{algpseudocode}
\usepackage{tabularray}
\usepackage{wrapfig}
\usepackage{makecell}
\usepackage{amsthm}

\definecolor{firstcolor}{HTML}{1cc61c}
\definecolor{secondcolor}{HTML}{cb0707}

\newtheorem{definition}{Definition}[section]
\newtheorem{corollary}{Corollary}[definition]

\title{Subgraph Generation for Generalizing on Out-of-Distribution Links}

\author{%
  Jay Revolinsky \\
  Michigan State University\\
  \texttt{revolins@msu.edu} \\
  \And
  Harry Shomer \\
  Michigan State University \\
  \texttt{shomerha@msu.edu} \\
  \AND
  Jiliang Tang\\
  Michigan State University \\
  \texttt{tangjili@msu.edu} \\
}

\begin{document}

\maketitle

\begin{abstract}
  Graphs Neural Networks (GNNs) demonstrate high-performance on the link prediction (LP) task. However, these models often rely on all dataset samples being drawn from the same distribution. In addition, graph generative models (GGMs) show a pronounced ability to generate novel output graphs. Despite this, GGM applications remain largely limited to domain-specific tasks. To bridge this gap, we propose FLEX as a GGM framework which leverages two mechanism: (1) structurally-conditioned graph generation, and (2) adversarial co-training between an auto-encoder and GNN. As such, FLEX ensures structural-alignment between sample distributions to enhance link-prediction performance in out-of-distribution (OOD) scenarios. Notably, FLEX does not require expert knowledge to function in different OOD scenarios. Numerous experiments are conducted in synthetic and real-world OOD settings to demonstrate FLEX's performance-enhancing ability, with further analysis for understanding the effects of graph data augmentation on link structures. The source code is available here \href{https://github.com/revolins/FlexOOD}{https://github.com/revolins/FlexOOD}.
\end{abstract}

\input{sections/intro}

\input{sections/preliminary}

\input{sections/flex}
\input{sections/experiments}

\input{sections/conclusion}



\begin{ack}
All authors are supported by the National Science Foundation (NSF) under grant numbers DGE2244164, CNS2321416, IIS2212032, IIS2212144, IOS2107215, DUE2234015, CNS2246050, DRL2405483 and IOS2035472, US Department of Commerce, Gates Foundation, the Michigan Department of Agriculture and Rural Development, Amazon, Meta, and SNAP.
\end{ack}

\bibliographystyle{unsrt}  
\small
\bibliography{Reference}
\normalsize


\appendix
\clearpage
\newpage


\input{sections/appendix}

\end{document}

%% file: sections/intro.tex
\section{Introduction} \label{sec:intro}

Graph Neural Networks (GNNs) demonstrate the ability to learn on graph data and have been used on a number of different downstream tasks that rely on understanding graph structure \cite{kipf2017semi}. Link Prediction (LP)\cite{liben2003link, li2024evaluating}, which attempts to predict unseen links in a graph, serves as one such example. For the task of LP, GNNs are used to learn node representations, which are then used to determine whether two nodes will form a link~\cite{kipf2016variational}. In recent years, advanced architectures have further enhanced state-of-the-art link prediction performance. To achieve this, the models often leverage important structural features directly within their neural architecture, allowing the model to effectively predict link formation \cite{wang2023neural, yun2021neo, shomer2024lpformer}. 

However, recent studies indicate that GNNs struggle to generalize to out-of-distribution (OOD) samples. This can arise when the underlying dataset properties are different during training and testing~\cite{gui2022good}. Additionally, this distribution shift for graphs is not well-aided by generalization techniques from other machine learning domains, such as CV and NLP \cite{li2022ood, gao2023alleviating}. Therefore, the study of the OOD problem has flourished for graph- and node-classification \cite{ ji2022drugood, koh2021wilds}. However, little direct attention has been paid to designing link prediction models that can better withstand such shifts in the underlying distribution~\cite{zhou2022ood, bevilacqua2021size}. This is an issue, as recent work~\cite{revolinsky2024understanding} has shown that current link prediction models struggle to generalize to shifts in the underlying structural distribution. Additionally, models that have been hypothesized to work are either too expensive or only function in a limited capacity~\cite{revolinsky2024understanding}.Given the success of out-of-distribution (OOD) generalization techniques in various graph-related tasks beyond link prediction~\cite{arjovsky2019invariant, krueger2021out, wu2024graph, wang2020tent}, a question arises regarding the relatively limited success of these methods in OOD link prediction problems. How can we improve out-of-distribution performance in link prediction?

\begin{figure*}[t]
    \centering
    \includegraphics[width=0.8\textwidth]{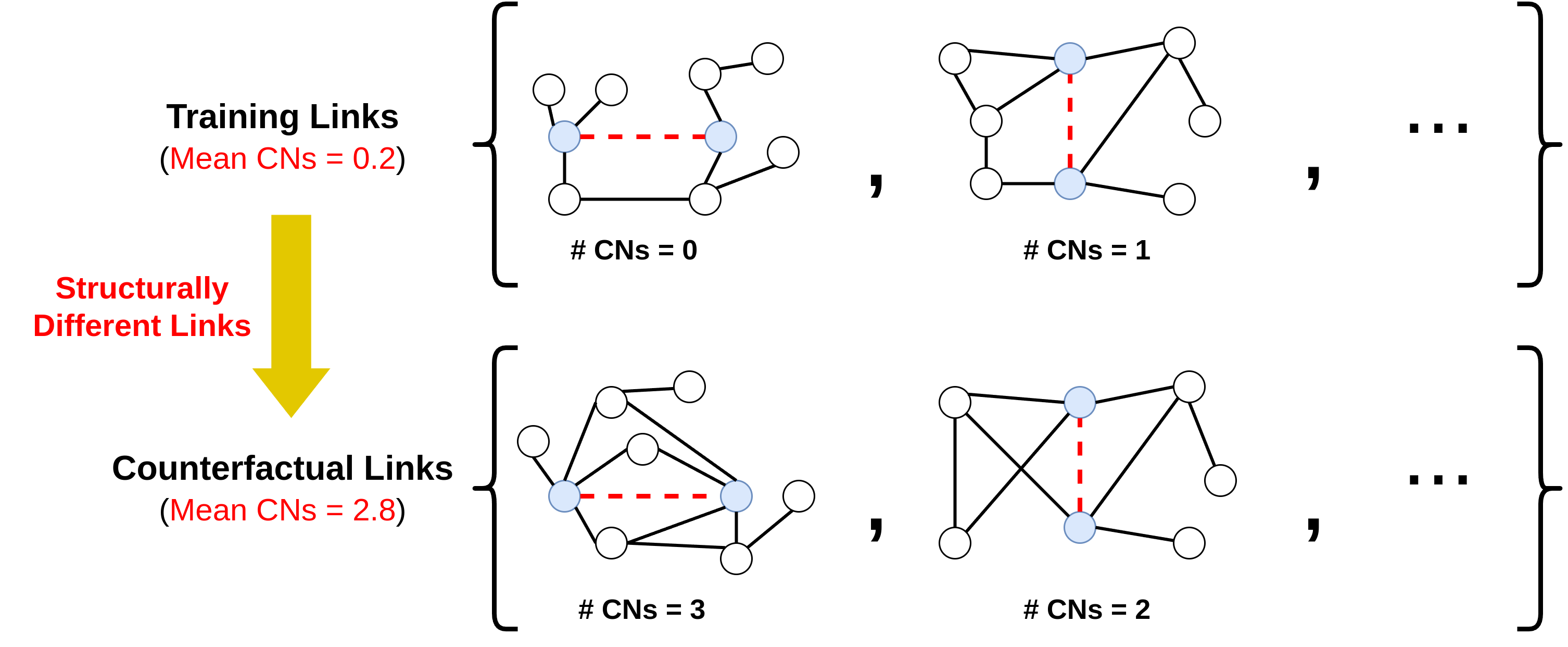}
    \caption{Example of counterfactual links that differ in terms of their structural properties such as Common Neighbors (CNs). In this example, the average training link typically contains very few CNs (0.2), thus we may want to generate counterfactuals with more CNs (2.8).}
    \label{fig:counter_example}

\end{figure*}

Intrinsically, out-of-distribution problems are difficult to manage; the simplest solution is to retrain or tune the model on new samples which are within distribution of the testing set \cite{bai2023feed}. Yet, this still means that the samples have to be acquired and tuned upon, or even detected that they fall out-of-distribution \cite{Wu_2023_ICCV, wu2023energy}. 
One such promising example occurs within both CV and NLP, where the training data is augmented with {\bf counterfactual samples}. Such counterfactual samples have been shown to be helpful for OOD tasks by improving the diversity of the training data problem~\cite{sun2022counterfactual}. The counterfactuals samples operate under the same causal rules as the original samples, even if the counterfactual was not originally contained within the training dataset \cite{ma2022clear}. An example for link prediction is shown in Figure~\ref{fig:counter_example}, where the counterfactuals are meant to be structurally different from the training samples. As shown, while the training samples have none or few common neighbors (i.e., shared 1-hop neighbors), the counterfactuals have multiple. The counterfactual links thus demonstrate an {\it alternative reason} for why some links may form. 
Within link prediction, counterfactuals have demonstrated the ability to enhance baseline model performance~\cite{pmlr-v162-zhao22e}. However, such techniques have been shown to be prohibitively expensive~\cite{revolinsky2024understanding}. Furthermore, they are reliant on non-learnable functions to generate counterfactuals, thus requiring prior knowledge of what type of shift is expected~\cite{pmlr-v162-zhao22e} which limits its real-world use.

Thus, an important question is, {\it how can we learn to efficiently generate new but meaningfully different samples to improve LP generalization?}
To address this issue, we apply graph generation as a  data augmentation method to generate samples which are {\it counterfactual} to the training distribution. The underlying principle behind this approach is to determine if it is possible to augment our training distribution to increase generalization and potentially improve LP performance. In order to achieve this, we design a new framework called {\bf FLEX} which leverages a generative graph model (GGM) co-trained with a GNN to produce subgraphs that are conditioned on a specific training link. The goal of the GGM is to take a single potential link (that is positive or negative) as input, and learn how to generate a new link that is counterfactually different in structure to the input. To ensure that the GGM learns to generate counterfactual links, a counterfactual loss is included to condition structural diversity and ensure we don't deviate too far from the original distribution. Furthermore, to avoid having to generate the entire adjacency for each new link, we instead propose to work directly with subgraphs, thus overcoming issues with efficiency.

Our contributions can be summarized as the following:
\begin{enumerate}
    \item Overall, we introduce {\bf FLEX}: a {\it simple yet effective} graph-generative framework that learns to generate counterfactual examples for improved link prediction performance.
    \item We demonstrate the effect of structural shifts through targeted analysis on link prediction model performance.
    \item We also conduct numerous experiments to show how FLEX can improve model generalization across multiple datasets and methods.
\end{enumerate}

%% file: sections/preliminary.tex
\section{Background and Related Work} \label{sec:preliminary}

We denote a graph as $\mathbf{G(X, A)}$, abbreviated to $\mathbf{G}$, where $\mathbf{X} \in\mathbb{R}^{n\times d}$ represents the node features in real space with $n$ nodes and feature dimensions $d$. $\mathbf{A} \in \{0,1\}^{n \times n}$ represents the adjacency matrix, within which nodes connect with one another to form edges, $e = (u,v)$. The $k$-hop subgraph of a node $v$ is denoted by $\mathbf{A}_{v}^{(k)}$. Consequently, the $k$-hop subgraph enclosed around a an edge $e$ is defined as $\mathbf{A}_{e}^{(k)} = \mathbf{A}_{u}^{(k)} \cup \mathbf{A}_{v}^{(k)}$.


{\bf Link Prediction}: Graph Neural Networks (GNNs)~\cite{kipf2017semi} are a common tool for modeling link prediction. GNNs learn representations relevant to graph structure as embeddings,  $\mathbf{H} = \text{GNN}(\mathbf{X}, \mathbf{A})$ which are then passed to link predictors to estimate whether a link will form or not.
However, several studies~\cite{zhang2021labeling, srinivasan2019equivalence} have shown that standard GNNs are not enough for link prediction as they ignore the pairwise information between two nodes. To account for this, recent methods inject such pairwise information into GNNs to elevate their capabilities. SEAL~\cite{zhang2018link} and NBFNet~\cite{zhu2021neural} consider message passing schemes that are conditional on a given link. To improve efficiency, other methods don't modify the message passing process, instead opting to include some link-specific information when scoring a prospective link. BUDDY applies a unique version of the labeling trick to subgraphs for generalizing on structural features \cite{chamberlain2022graph}. NCN/NCNC~\cite{wang2023neural} and Neo-GNN~\cite{yun2021neo} both elevate traditional link heuristics via neural operators to better understand link formation. Lastly, \cite{shomer2024lpformer} proposes a more general scheme for estimating the pairwise information between nodes that adaptively learns how two nodes relate. A core component of these models is their increased reliance on the substructures contained within the graph datasets, which improves the model's expressivity but can affect prediction performance in OOD scenarios~\citep{mao2024demystifying}.

{\bf Graph Generative Models}: We treat graph generation as output of a scoring function \( s : \mathbb{R}^d \times \mathbb{R}^d \rightarrow \mathbb{R} \) for quantifying similarity between node embeddings, which is often defined as an inner product: $s(u, v) = \mathbf{H}_{u}^\top \mathbf{H}_{v}$ and further calculated as edge-probabilities, $P((u,v) \in E \mid \mathbf{H}_{u}, \mathbf{H}_{v}) = \sigma(s(i, j))$, where \( \sigma(\cdot) \) is the sigmoid function. Whereas, we focus on the capability of auto-encoders inferring from latent embeddings to re-produce an adjacency matrix \cite{kipf2016variational}. More advanced graph generation models exist: such as auto-regressive, diffusion, normalizing-flow, and generative-adversarial networks \citep{you2018graphrnn, vignac2022digress, luo2021graphdf, martinkus2022spectre}. However, these models often employ mechanisms which restrict their applications beyond graph generation. For example, discrete-denoising models such as DiGress generate a new adjacency matrix with discrete space edits, which can be computationally restrictive to re-train when generalizing on a variety of different graph structures~\cite{kong2023autoregressive}.

{\bf Methods for OOD}: Numerous methods exist to improve the generalization performance of neural models~\citep{arjovsky2019invariant, krueger2021out, sun2016deep, sagawa2019distributionally, ganin2016domain}. Most of these methods operate underneath the invariance learning principle, which divides training data into environmental subsets and then conditions the model to better identify any variance within the training subsets. However, these methods have shown that careful considerations are required for effective improvement of OOD performance~\cite{gulrajani2020search}. Additionally, it has also been shown that generalizing with these techniques is difficult to apply properly within node- and graph-classification~\cite{li2022learning}. Therefore, architectures and techniques which elevate invariance principles to be better-suited for graph data are employed to improve GNN performance~\cite{chen2023does, zhang2022dynamic}. Recently, graph generation has been applied within OOD scenarios as well. For example, EERM is a technique which integrates graph generators within a model to improve OOD performance on graphs. However, the generators can lead to scalability issues when considering additional nodes for link formation~\cite{wu2022eerm}. GOLD leverages latent generative models to learn on OOD samples, yet it functions predominantly for OOD detection on graphs and not directly improving OOD generalization in link prediction~\cite{wang2025gold}. Lastly, CFLP~\cite{pmlr-v162-zhao22e} considers generating counterfactual links for enhancing link prediction. However, their proposed algorithm is (a) a non-parametric method that relies on the Louvain~\cite{blondel2008fast} algorithm, (b) has been shown to be prohibitive to run, even on smaller datasets~\cite{revolinsky2024understanding}.

%% file: sections/flex.tex
\section{FLEX} \label{sec:flex_model}

In Section~\ref{sec:intro}, we introduced the OOD problem for link prediction and how graph generation can serve as a means to solve the problem. That is, we can leverage graph generation to learn to generate counterfactual links, with substructures different from our training links. However, {\it is it possible to generate such counterfactual links?} Effectively, there are endless ``meaningless'' graphs with no relevant structure to an original dataset; which are also likely detrimental to downstream model performance. Therefore, applying graph data augmentation to improve performance requires understanding of the structure within the graph dataset \cite{singh2021edge}. It's thus desirable to build a learnable framework which understands the formation of links and can target aspects of the data that are most relevant to performance (like structural features). To achieve this, we introduce {\bf FLEX}, the \textbf{F}ramework for \textbf{L}earning to \textbf{EX}trapolate Structures in Link Prediction. As a graph data augmentation framework, FLEX utilizes a variety of techniques to ensure: computability, scalability, and expressiveness.

\begin{figure*}[t]
    \centering
    \includegraphics[width=0.85\textwidth]{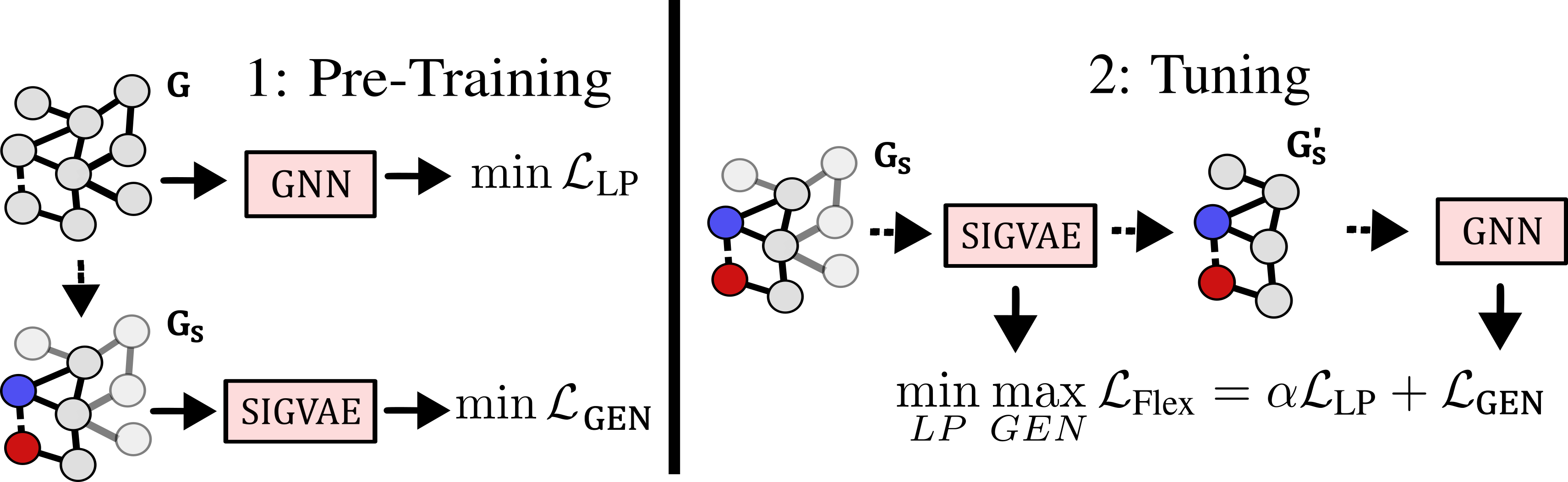}
    \caption{An illustration of the FLEX framework for a single dataset sample. {\bf Step 1} involves pre-training both models separately to optimize their performance, like in real-world scenarios. {\bf Step 2} involves adversarial co-training of the two models, where the GGM generates synthetic samples to tune the GNN.}
    \label{fig:flex_frame}
    \vskip -1em
\end{figure*}

Following these principles, FLEX then functions in two critical steps, as illustrated in Fig.~\ref{fig:flex_frame}. {\bf First}, we pre-train a GNN on the dataset's full adjacency matrix by optimizing the predictive loss, $\mathcal{L}_{\text{LP}}$. The intent of pre-training the GNN serves as a means of simulating a real-world scenario, where we may only wish to improve a pre-existing model's ability to generalize on OOD samples \cite{gui2022good, krueger2021out}. A graph generative model (GGM) is then pre-trained separately to minimize it's generative loss, $\mathcal{L}_{\text{GEN}}$. The GGM is conditioned on each sample (i.e., link) via the labeling trick on the \textit{k}-hop enclosed subgraph~\cite{zhang2021labeling}. This ensures that we can generate a {\it new link} that is counterfactual to an {\it existing link}. {\bf Second}, we apply both pre-trained models in a co-training framework, where the GGM produces synthetic dataset samples as input for fine-tuning the GNN. The GGM maximizes the distance between posterior and prior while the GNN attempts to minimize it's prediction loss; much like adversarial-conditioning in GANs and other auto-encoder frameworks \cite{goodfellow2020generative, yang2019conditional, wang2025gold}. As such, the GNN prediction loss functions as a means of retaining enough information from the original dataset distribution to act as efficient counterfactual conditioning to improve OOD performance.

\subsection{General Motivation} \label{sec:motivation}

The main objective of the FLEX framework is to generate graph samples which retain node feature properties but also produce edge structures which are counterfactual to the original data. This is feasible with any type of well-trained graph generative model (e.g., auto-encoders~\cite{kipf2016variational} or diffusion models~\cite{vignac2022digress}). To produce relevant counterfactuals for link prediction, we consider the following definitions.

\begin{definition}[Expressive Link Features] \label{def:link_feats}
Consider an edge sample $e = (u, v)$, and it's $k$-hop subgraph $\mathbf{A}_{e}^{(k)}$. We want to learn an encoder $f_{\theta}(\cdot)$ that can operate on $\mathbf{A}_{e}^{(k)}$ and learn to extract structural features $\mathbf{H}_e$ that are specific to the link $(u, v)$ (e.g., link heuristics~\cite{newman2001clustering, katz1953new}). We assume that $f_{\theta}(\cdot)$ is expressive such that it can extract link-specific features. We then represent the probability distribution of the features extracted by the encoder to be $\mathbb{P}_H(\mathbf{A}_{e}^{(k)}) = f_{\theta}(\mathbf{A}_{e}^{(k)})$. 
\end{definition}

In order to properly generate counterfactual links, we need an understanding of {\it what} our generated samples should be counterfactual to. Intuitively, this should be link properties that are known to be relevant and important for link prediction. As shown in the next definition, an encoder $f_{\theta}(\cdot)$ that can extract expressive link features is therefore necessary for producing proper counterfactual links. As if $f_{\theta}(\cdot)$ is not suitably expressive, our generative model will be unable to distinguish link structure and fail to generate counterfactuals relevant to link prediction.

\begin{definition}[Counterfactual Structure] \label{def:counter}
For an edge sample $e = (u, v)$, a meaningfully different sample (counterfactual) -- $\mathbf{\tilde{A}}_{e}^{(k)}$ exists where the link feature distribution estimated between the original subgraph and it's counterfactual are approximately non-equivalent, $\mathbb{P}_H(\mathbf{A}_{e}^k) \not\approx \mathbb{P}_H(\mathbf{\tilde{A}}_{e}^{(k)})$.
\end{definition}

A proper counterfactual sample should have different underlying link features from the original sample.  
An example is given in Figure~\ref{fig:counter_example} where we assume that we have an encoder that can extract common neighbors (CNs)~\cite{newman2001clustering}, a vital link heuristic. Since the training samples have no or few CNs, the corresponding counterfactuals will then have several CNs. These new samples are thus {\it meaningfully different}, in that they differ in an important and realistic feature for link prediction.

\begin{corollary}[Feature-Conditional Equivalence]
Given the previous definition of counterfactual structure, the link features contained within $k$-hop subgraph  $\mathbf{A}_e^{(k)}$ are not invariant in isolation as we must consider the node features. Therefore, in order for $\mathbf{\tilde{A}}_{e}^{(k)}$ to maintain a valid counterfactual structure, it must be conditioned on the node features $\mathbf{X}_{e}^k$ within the original subgraph. 
That is, $\mathbb{P}_H(\mathbf{A}_{e}^{(k)} \mid \mathbf{X}_{e}^k) = f_{\theta}(\mathbf{A}_{e}^{(k)} \mid \mathbf{X}_{e}^k))$ and $\mathbb{P}_H(\mathbf{\tilde{A}}_{e}^{(k)} \mid \mathbf{X}_{e}^k) = f_{\theta}(\mathbf{\tilde{A}}_{e}^{(k)} \mid \mathbf{X}_{e}^k))$. For convenience, we further write this as $\mathbb{P}_H(\mathbf{G}_{e}^{(k)}) = f_{\theta}(\mathbf{G}_{e}^{(k)})$ and $\mathbb{P}_H(\mathbf{\tilde{G}}_{e}^{(k)}) = f_{\theta}(\mathbf{\tilde{G}}_{e}^{(k)})$.
\end{corollary}

Therefore, the counterfactual structure is dependent on the compatibility between $\mathbf{\tilde{A}}$ and $\mathbf{X}$. A failure to properly condition structure on $\mathbf{X}$ will not fulfill the definition for a counterfactual structure since the generated node features will indicate spurious distributional mismatch relative to the original subgraph samples. So, the encoder $f_{\theta}(\cdot)$ must also consider the original node features as input.

Given these definitions, we can see that to generate proper counterfactuals we first need to extract expressive link features that are conditional on the node features. Then, we need to learn a Generative Graph Model (GGM) which takes said features to output a new sample whose features follow a different distribution. In order to do this, we must ensure three things: \textit{(a) Scalability:}  In order to ensure relevance to real-world problems, the GGMs must operate on large graphs. \textit{(b) Expressiveness:} First, the extracted features for each link must be suitably expressive. Second, the GGM itself will need to effectively sample from complicated distributions to produce relevant graph structures. \textit{(c) Counterfactual:} Generated structures must indicate a level of change which does not replicate the training distribution but retains meaningful correlation. In the rest of this section we outline our method for tackling these challenges.

\subsection{Semi-Implicit Variation for Out-of-Distribution Generation}

Following principle (a.) from Section~\ref{sec:motivation}, the scalability of the practical implementation becomes a concern. Computational complexity of more refined GGMs can be restrictive, whereas less computationally-intensive generative models may result in low-quality generations \cite{simonovsky2018graphvae, yanswingnn}. To balance this, we employ semi-implicit variation~\cite{yin2018semi}, for it's inherent scalability when implemented in an auto-encoder and it's expressiveness for modeling complex distributions. 


Let the true data-generating distribution be $p(G)$, and assume it is modeled via a latent variable model with latent code $H$ and a semi-implicit posterior of the form: 
\begin{equation} \label{equation:1}
q_{\phi}(H_{e} \mid {\tilde{X}}_{e}^{(k)}, {\tilde{A}}_{e}^{(k)}) = \int q_{\phi}(H_{e} \mid \psi) \, q_{\phi}(\psi \mid {X}_{e}^{(k)}, {\tilde{A}}_{e}^{(k)}) \, d\psi,
\end{equation}
where $q_{\phi}(\psi \mid X, A)$ is a flexible (potentially implicit) distribution. Suppose the model is trained to maximize the semi-implicit evidence lower bound (ELBO)~\cite{hasanzadeh2019semi}:
\begin{equation} \label{equation:2}
\mathcal{L}_{\text{SIVI}} = \mathbb{E}_{\psi \sim q_{\phi}(\psi \mid {X}_{e}^{(k)},{A}_{e}^{(k)})}\left[ \mathbb{E}_{H \sim q_{\phi}(H \mid \psi)} \left[ \log p({A}_{e}^{(k)} \mid H_{e}) \right] - \mathrm{KL}(q_{\phi}(H_{e} \mid \psi) \parallel p(H_{e})) \right],
\end{equation}
and assume $p(\mathbf{H}_{e})$ is a broad prior (e.g., isotropic Gaussian), and $p(\mathbf{A_{e}} \mid \mathbf{H}_{e})$ defines a valid graph decoder. Then, given an auto-encoder with an expressive architecture capable of distinguishing the structure within samples drawn from $q_{\phi}$ and $p$, sampling from $
\mathbf{H_{e}} \sim q_{\phi}(\mathbf{H}_{e} \mid \psi), \quad\psi \sim q_{\phi}(\psi)$ yields synthetic graphs $\mathbf{\tilde{G}_{e}} = (\mathbf{X}_{e}, \mathbf{\tilde{A}_{e}})$ whose features are derived from the original dataset distribution but reveal emergent out-of-distribution (OOD) structure with respect to the training data $\mathcal{D}_{\text{train}} \sim \mathbb{P}(\mathbf{G})$, provided that $q_{\phi}(\psi) \not\approx q_{\phi}(\psi \mid \mathcal{D}_{\text{train}})$. That is, the complete generative process follows:
\begin{equation} \label{equation:3}
\mathbf{\tilde{G}_{e}} \sim p_{\theta}(\mathbf{\tilde{G}}_{e} \mid \mathbf{H}_{e}), \quad \mathbf{H}_{e} \sim q_{\phi}(\mathbf{H}_{e} \mid \psi), \ \psi \sim q_{\phi}(\psi),
\end{equation}
Therefore, Eq.~\eqref{equation:3} defines a valid procedure for generating OOD graph samples. In scenarios where the sampled distribution is not a broad prior, this process then decomposes further to a standard variational generative process \cite{hasanzadeh2019semi, kipf2016variational}. 

As a learnable mechanism, semi-implicit variance ($\psi)$ often relies on inputting randomness into prior distributions; this randomness can then be treated as an adversarial noise, much like how OOD samples would appear to pre-trained GGMs. As such, an auto-encoder which effectively models semi-implicit variance of training distributions can generate complicated graph samples which mimic input structure, fulfilling the expressiveness principle while maintaining the scalability of an auto-encoder~\cite{hasanzadeh2019semi, simonovsky2018graphvae}. We show in Section~\ref{ssec:ablation} that the use of a semi-implicit GGM or a standard graph GGM is necessary for strong counterfactual generation.


\subsection{Link-Specific Subgraph Generation}

Semi-implicit variation assumes that a GGM can learn to generate $\mathbf{\tilde{G}_e}$. However, as noted in Definition~\ref{def:link_feats}, to make this task relevant to link-prediction and continue fulfilling the expressiveness principle, we must first learn to extract {\it link-specific features}. That is, we want an encoder $f_{\theta}(G_e^{(k)})$ that can extract such features from the k-hop neighborhood of a link $e=(u, v)$.  Only then will our GGM have the suitable amount of information to generate meaningful counterfactuals that differ in key link properties.  

To achieve this, the encoder $f_{\theta}(\cdot)$ should be able to effectively encode the graph conditional on a specific link. The link-specific representations will then be used by the GGM for generation. \citep{zhang2021labeling} show that standard GNNs aren't expressive to links. To combat this, they introduce the labeling trick that ensures that a given GNN can learn to distinguish target links from other nodes within a graph sample. They demonstrate that the labeling trick can extract a number of different relevant structural features for a link~\cite{zhang2018link}. 

The labeling trick is defined as a function $\ell : \mathbf{A}^{(k)} \to \{0, 1\}$ where for a link $e=(u, v)$ the value for a sampled node $x$ is given by:
\begin{equation} \label{equation:4}
\ell(x) = 
\begin{cases}
1, & \text{if } x = u \textit{ or } x = v \\
0, & \text{else } \\
\end{cases}
\end{equation}
This results in a labelled subgraph $L_{e}^{(k)}$ which is fed, along with the node features, to a GNN to produce the link-specific representations:
\begin{equation}
    \mathbf{H}_{e} = \text{GNN}(L_{e}^{(k)}, X_e^{(k)}). 
\end{equation}
Given that all edges within a graph are viable link prediction targets, an effective zero-one labeling requires extracting the $k$-hop enclosed subgraphs conditioned on a target edge, $\mathbf{G}_{e}^{(k)}$. When these subgraphs are restricted to a smaller size, this reduces the direct computation required from the GGM to model subgraph distributions, ensuring FLEX's scalability principle~\cite{zhang2018link}.

\subsubsection{Node-Aware Decoder} \label{sec:node_decode}

Furthermore, to continue ensuring scalability and expressiveness. The decoder for FLEX's GGM is made aware of the independent number of nodes within subgraph samples for a given mini-batch along the block diagonal matrix, $A=\mathrm{diag}(A_1, \dots, A_K)$ with $A_i \in \mathbb{R}^{\mathcal{N}_i \times \mathcal{N}_i}$. This ensures that generated subgraphs retain the original number of input nodes and prevent message-passing along edges between distinct subgraph samples.

Within early experiments, as shown in Figure~\ref{fig:0ThreshCora}, generated subgraph samples suffered from the degree-bias phenomenon~\cite{tang2020investigating}. Wherein, the backbone GNN learns on nodes with a higher number of edges at a much-greater frequency than low-degree nodes, prioritizing learning information from the high-degree nodes~\cite{liu2023generalized}. Therefore, the downstream generation results in subgraphs where the average node had a high degree and are thus very dense, regardless of the node-degree within the input graph. We verify this phenomenon in Appendix~\ref{sec:app_deg}. To account for this, we apply an indicator function to FLEX-generated subgraphs which eliminates edges with lower probability than a threshold, $\gamma$:
\begin{equation} \label{equation:5}
\tilde{p}(u,v) = p(u,v) \cdot \mathbb{I}[p(u,v) \geq \gamma].
\end{equation}
This function only keeps those links that have a high probability, thus constraining the generative model to choose the links which it is most confident in.
As such, the indicator function prevents densely-connected graphs, especially for OOD scenarios where training on dense graphs may not be desirable for downstream performance. The value of the threshold $\gamma$ is treated as a hyperparameter. In Section~\ref{ssec:ablation}, we show how the value of $\gamma$ impacts performance.

\subsection{Generating Counterfactual Links}

As part of FLEX, all previous components work to produce meaningful subgraphs. However, it is still necessary for the GGM to learn how to produce subgraph samples which are structurally-dissimilar from training, while retaining relevance to the node features within the training distribution. 

As discussed in Definition~\ref{def:counter}, in order to ensure that the generated links are structurally-dissimilar (i.e., counterfactual) we want to generate links whose structural feature distribution is different from a target input link. That is, for an input training sample $e=(u, v)$ and it's counterfactual, we want that $\mathbb{P}_H(\mathbf{A}_{e}^k) \not\approx \mathbb{P}_H(\mathbf{\tilde{A}}_{e}^{(k)})$ where $\mathbf{\tilde{A}}_{e}^{(k)} = p_{\theta}(\mathbf{\tilde{G}_e} \mid \mathbf{H}_e)$. Therefore, to achieve this goal of having the GGM produce different structures, we consider {\it maximizing the difference in structure}. That is, we optimize the GGM to maximize the difference in input and generated samples; $\max \mathcal{L}_{\text{GEN}}$ where $\mathcal{L}_{\text{GEN}}$ is defined as in Eq.~\eqref{equation:1}. By doing so, we ensure that the newly generated samples will indeed be structurally different from the input training samples. 

However, just blindly maximizing the generative loss will result in generated subgraphs that are structurally incoherent and meaningless to our training samples. In reality, we really want the generated distribution to modestly differ in key structural features. We ensure this in two ways. First, we apply a quadratic penalty to the generative loss $\mathcal{L}_{\text{GEN}}$. The penalty is centered around a target value, $\tau$. This penalty restricts any shifts to the posterior distribution, meaning generated graphs will only deviate slowly from the prior distribution and prevent the samples from devolving into noise. This is given by the following,
\begin{equation} \label{equation:6}
\mathcal{L}_{\text{GEN}} = \mathcal{L}_{\text{SIVI}} -\left( \mathrm{KL} \left( \mathbb{E}_{\psi \sim q_\phi(\psi \mid X_{e}, A_{e})} \left[ q(H_{e} \mid \psi) \right] \,\big\|\, p(H_{e}) \right) - \tau \right)^2.
\end{equation}
Second, we also attempt to correctly classify the link based on it's original label. That is, we want to predict the existence of the original link based on the newly generated sample. This serves as a means for inducing learnable counterfactual treatment within the GGM. If the generative model deviates too far from the training distribution or considers useless structural features, the GNN will be unable to cope, thus resulting in poor classification performance. It therefore allows for a ``check'' on the generation quality, limiting the potential for incoherent generation.

The final optimization goal of FLEX is given by the following, $\mathcal{L}_{\text{LP}}$ denotes the classification loss (BCE): 
\begin{equation} \label{equation:7}
\min_{LP}\max_{GEN}\mathcal{L}_{\text{Flex}} =  \alpha\mathcal{L}_{\text{LP}} + \mathcal{L}_{\text{GEN}}
\end{equation}
$\alpha$ represents the weight assigned to the counterfactual predictions produced by the GNN tuned within the FLEX framework. Since the co-trained GNN is tuned on synthetic samples, the minimization of $\mathcal{L}_{\text{LP}}$ ensures that the GNN retains it's ability to predict on positive and negative samples while also conditioning the maximization of $\mathcal{L}_{\text{GEN}}$. In tandem, the two function in an adversarial co-optimization to predict on samples with increasingly different structures~\citep{pan2018adversarially, wang2025gold}.  

We further illustrate the overall framework in Figure~\ref{fig:flex_frame}.  In the first stage both the GNN and GGM are trained separately. Then in the second stage, the components are co-trained via the objective defined in Eq.~\eqref{equation:7}. Both procedures are described further in Algorithm~\ref{alg:FLEX}. Thus, FLEX is a lightweight and straightforward method for enhancing OOD performance for link prediction. In the next section, we test FLEX, showing it's ability to improve OOD performance for link prediction.

%% file: sections/experiments.tex
\section{Experiments} \label{sec:experiments}
We now evaluate FLEX to answer the following research questions. \textbf{RQ1:} Does FLEX contribute to better link prediction performance in OOD scenarios? \textbf{RQ2:} How might separate components of the FLEX framework improve OOD performance? \textbf{RQ3:} How sensitive is FLEX to different hyperparameter settings? \textbf{RQ4:} Does FLEX learn to generate samples with counterfactual structure?

\subsection{Setup}  The benchmarking experiments apply two different GNN backbones, Graph Convolutional Network (GCN) and Neural Common Neighbor (NCN) \cite{kipf2017semi, wang2023neural}. We then compare against the well-known OOD and generalization methods as baselines: CORAL, DANN, GroupDRO, VREx, IRM~\cite{sun2016deep, ganin2016domain, sagawa2019distributionally, krueger2021out, arjovsky2019invariant}. Note that CFLP~\cite{pmlr-v162-zhao22e} is omitted as we were unable to scale it to ogbl-collab and most LPShift datasets. Detailed hyperparameter settings are included within Appendix~\ref{sec:hyper}. For datasets, we consider the synthetic datasets generated via the protocol designed by LPShift~\cite{revolinsky2024understanding}. Please see Appendix~\ref{sec:synth} for more details. We also test on the original ogbl-collab split~\cite{hu2020open} which has been shown to contain a distribution shift in CNs~\cite{zhu2021neural}. Lastly, all synthetic datasets are evaluated using Hits@20 while ogbl-collab is evaluated with Hits@50.

\subsection{RQ1: Flex Performance} \label{ssec:perf}

 We evaluate FLEX on six unique datasets and seven dataset splits, with one real-world temporal split for ogbl-collab and an additional six different synthetic dataset splits. As shown in Table~\ref{table:flex_full_results}, FLEX improves the performance for 26 out of 27 datasets when applied to GCN, and for all 27 datasets when applied to NCN. This leads to aggregate average increase of {\bf 5.13\% for GCN and 28.36\% for NCN}. On the other hand, other baselines either perform worse or on-par with GCN. This indicates that FLEX generates subgraphs which can improve model generalization.

\begin{table}[h]
\centering
\caption{Results for the LPShift Datasets by direction ({\it forward} or {\it backwards}) and type ({\it CN, SP} or {\it PA}). Ordered from bottom up: Collab, PubMed, Cora, CiteSeer, PPA. \textit{Note:} PPA for PA and SP is missing due to taking >24h. Results for the original ogbl-collab~\cite{hu2020open} are  included as {\it real}. We highlight in \textcolor{blue}{blue} when FLEX increases over the base model and \textcolor{red}{red} otherwise.}
\begin{adjustbox}{width=1\textwidth}
\begin{tabular}{cc|c|c|c|c|c|c|c|c|c}

\toprule
\multicolumn{2}{c|}{\multirow{2}{*}{Dataset}} & \multicolumn{9}{c}{Models} \\ 
&& CORAL & DANN & GroupDRO & VREx & IRM & GCN & GCN+FLEX & NCN & NCN+FLEX \\ 
\midrule

\multirow{14}{*}{\rotatebox{90}{Forward}} & \multirow{6}{*}{CN} &30.93 ± 0.24 & 30.86 ± 0.32&27.83 ± 1.76 & 30.93 ± 0.24&25.78 ± 2.04 &31.92 ± 0.25 &\textcolor{blue}{32.87 ± 0.23} &1.62 ± 5.04&\textcolor{blue}{3.95 ± 0.75} \\
& &67.75 ± 2.49 &68.11 ± 3.04 &65.27 ± 3.50 &66.54 ± 2.42 &66.67 ± 1.50 &67.18 ±  2.43 &\textcolor{blue}{68.24 ± 1.30} &75.83 ± 4.42&\textcolor{blue}{79.34 ± 0.1}\\
& &57.45 ± 1.70 &57.54 ± 2.80 &38.21 ± 5.63 &53.15 ± 3.58 &55.30 ± 2.54 &56.22 ± 1.31 &\textcolor{blue}{57.78 ± 0.08} &75.91 ± 1.50&\textcolor{blue}{79.34 ± 0.10}\\
& &71.32 ± 0.32 &71.62 ± 0.42 &57.25 ± 1.95 &71.60 ± 0.66 &68.18 ± 0.48 &69.60 ± 0.45 &\textcolor{blue}{70.04 ± 0.01} &96.63 ± 0.24&\textcolor{blue}{96.72 ± 0.32}\\
& &42.60 ± 1.61 &43.60 ± 1.21 &22.71 ± 4.03 &43.61 ± 1.21 &42.04 ± 1.32 &43.14 ± 1.22 &\textcolor{blue}{43.23 ± 0.05} &2.37 ± 0.02&\textcolor{blue}{3.39 ± 0.09}\\
\cmidrule{2-11}
& \multirow{4}{*}{PA} &69.85 ± 3.79 &67.57 ± 4.72 &51.80 ± 7.12 &69.03 ± 2.92 &68.28 ± 3.63 &68.88 ± 3.34 &\textcolor{blue}{70.83 ± 0.41}&65.64 ± 1.27&\textcolor{blue}{67.65 ± 0.26}\\
& &52.39 ± 4.16 &49.24 ± 6.44 &40.16 ± 6.56 &51.05 ± 3.63 &50.03 ± 3.06 &55.13 ± 5.30 &\textcolor{blue}{56.58 ± 5.22} &53.44 ± 1.52&\textcolor{blue}{53.59 ± 0.08}\\
& &83.35 ± 0.65 &83.19 ± 0.50 &66.00 ± 3.71 &81.43 ± 0.80 &75.68 ± 0.81 &82.04 ± 0.95 &\textcolor{blue}{84.09 ± 0.65} &88.35 ± 0.19&\textcolor{blue}{88.71 ± 0.11}\\
& &61.39 ± 1.19 &61.69 ± 1.33 &39.92 ± 5.11 &61.52 ± 0.92 &60.27 ± 0.66 &63.83 ± 1.04 &\textcolor{blue}{63.93 ± 1.20} &65.66 ± 0.50&\textcolor{blue}{65.94 ± 0.32}\\
\cmidrule{2-11}
& \multirow{4}{*}{SP} &42.35 ± 1.52 &35.53 ± 5.14 &30.69 ± 2.43 &44.60 ± 2.57 &39.18 ± 3.79 &44.60 ± 2.57 &\textcolor{blue}{45.85 ± 0.24} &52.06 ± 2.99&\textcolor{blue}{54.21 ± 0.36}\\
& &26.26 ± 3.22 &26.89 ± 3.62 &19.63 ± 2.65 &25.91 ± 2.88 &24.13 ± 4.01 &24.82 ± 3.40 &\textcolor{blue}{29.91 ± 0.19} &48.31 ± 1.91&\textcolor{blue}{49.68 ± 0.08}\\
& &67.41 ± 2.15 &68.03 ± 1.03 &51.49 ± 3.49 &68.18 ± 1.63 &64.28 ± 1.98 &68.52 ± 1.29 &\textcolor{blue}{69.24 ± 1.19} &77.91 ± 0.48&\textcolor{blue}{79.10 ± 0.02}\\
& &40.36 ± 1.86 &39.07 ± 2.43 &32.82 ± 2.54 &40.45 ± 2.35 &38.63 ± 2.28 &39.13 ± 2.16 &\textcolor{blue}{41.22 ± 3.55} &8.23 ± 2.60&\textcolor{blue}{26.83 ± 23.95}\\
\midrule
\multirow{14}{*}{\rotatebox{90}{Backward}} & \multirow{6}{*}{CN} & 13.52 ± 1.01&14.31 ± 0.49 & 11.70 ± 0.81&13.46 ± 1.17 & 11.34 ± 2.84&14.19 ± 0.46 &\textcolor{blue}{14.49 ± 0.51}  &1.21 ± 0.53&\textcolor{blue}{2.62 ± 0.14}\\
& &41.88 ± 4.38 &42.37 ± 4.62 &31.78 ± 6.07 &41.27 ± 6.01 &41.83 ± 3.25 &41.03 ± 5.68 & \textcolor{blue}{43.96 ± 1.18}&34.70 ± 4.12&\textcolor{blue}{38.65 ± 0.18}\\
& &43.13 ± 5.13 &40.72 ± 3.60 &26.36 ± 3.19 &40.68 ± 2.76 &38.60 ± 3.79 &39.92 ± 1.09 &\textcolor{blue}{44.87 ± 0.32} &45.04 ± 2.57&\textcolor{blue}{46.32 ± 1.02}\\ & &28.96 ± 0.77 &26.77 ± 0.51 &15.57 ± 2.02 &27.91 ± 0.41 &27.24 ± 0.67 &28.67 ± 0.57 &\textcolor{blue}{29.31 ± 0.12} &22.16 ± 0.66&\textcolor{blue}{22.43 ± 0.03}\\
& &24.16 ± 0.72 & 25.07 ± 0.67&21.03 ± 1.37 &24.40 ± 0.51 &21.86 ± 0.64 &24.62 ± 0.73 &\textcolor{blue}{25.24 ± 0.01} &7.18 ± 0.42&\textcolor{blue}{11.62 ± 5.27}\\
\cmidrule{2-11}
& \multirow{4}{*}{PA} &38.68 ± 3.39 &38.13 ± 3.52 &16.16 ± 7.56 &38.33 ± 2.19 &31.26 ± 4.09 & 37.67 ± 2.87&\textcolor{blue}{39.70 ± 0.26}&35.30 ± 2.55&\textcolor{blue}{39.49 ± 0.22}\\
& &38.90 ± 1.79 &38.45 ± 3.22 &25.10 ± 2.32 &37.63 ± 1.87 &37.88 ± 1.11 &38.00 ± 1.24 &\textcolor{blue}{40.07 ± 0.14} &24.69 ± 5.02&\textcolor{blue}{26.63 ± 0.10}\\
& &26.86 ± 0.97 &27.51 ± 0.56 &19.38 ± 4.85 &28.40 ± 0.74 &25.25 ± 2.95 &29.04 ± 1.58 &\textcolor{blue}{35.94 ± 4.60} &22.10 ± 3.30&\textcolor{blue}{27.05 ± 0.09}\\
& &72.45 ± 0.71 &72.68 ± 0.82 &9.77 ± 2.30 &72.45 ± 0.30 &54.50 ± 3.72 &73.38 ± 0.94 &\textcolor{red}{59.58 ± 1.12} &70.66 ± 5.33&\textcolor{blue}{72.04 ± 0.02}\\
\cmidrule{2-11}
& \multirow{4}{*}{SP} &19.30 ± 4.72 &16.51 ± 6.82 &11.51 ± 3.65 &16.98 ± 5.12 &15.81 ± 2.58 & 16.98 ± 5.12 &\textcolor{blue}{22.09 ± 1.10} &23.95 ± 4.32&\textcolor{blue}{41.63 ± 0.49}\\
& &24.65 ± 3.66 &27.02 ± 3.20 &17.81 ± 3.92 &26.67 ± 3.49 &25.96 ± 3.86 &26.67 ± 3.49 &\textcolor{blue}{28.25 ± 1.23} &22.81 ± 2.77&\textcolor{blue}{24.30 ± 0.93}\\
& &22.39 ± 2.29 &23.05 ± 1.80 &10.59 ± 3.42 &22.61 ± 1.73 &20.92 ± 2.44 &22.61 ± 1.73 &\textcolor{blue}{24.93 ± 0.23} &23.82 ± 1.54&\textcolor{blue}{25.44 ± 0.34}\\
& &33.50 ± 0.57 &33.94 ± 0.40 &33.40 ± 0.94 &33.48 ± 0.58 &32.97 ± 0.57 &33.58 ± 0.47 &\textcolor{blue}{33.62 ± 0.49} &19.87 ± 1.02&\textcolor{blue}{20.99 ± 1.67}\\
\midrule
Real & Collab &49.49 ± 0.86 &48.48 ± 1.78 &44.30 ± 0.61 &49.35 ± 0.75 &46.26 ± 1.09 &50.40 ± 1.01 &\textcolor{blue}{52.42 ± 0.08} &64.83 ± 0.18&\textcolor{blue}{64.99 ± 0.32}\\
\midrule
 & Avg ($\Delta$\%) & \textcolor{blue}{+0.06} & \textcolor{red}{-1.03} & \textcolor{red}{-29.31} & \textcolor{red}{-0.78} & \textcolor{red}{-7.14} &--&\textcolor{blue}{\bf+5.13} &-- & \textcolor{blue}{\bf+28.36}\\
\bottomrule
\end{tabular}
\label{table:flex_full_results}
\end{adjustbox}
\vskip -0.5em
\end{table}

\begin{wraptable}{r}{0.4\textwidth}
\centering
\caption{Ablation across the LPShift "Backwards" CN Splits.}
\begin{adjustbox}{width=0.4\textwidth}
\begin{tabular}{c|c|c|c|c}
\toprule
\multicolumn{1}{c|}{Dataset} & \multicolumn{3}{c}{Models} \\ 
& \textbf{FLEX} & \textbf{w/o SEAL} & \textbf{w/o LP Loss} & \textbf{w/o SIGVAE} \\ 
\midrule
Cora &{\bf 44.87} ± 0.32&34.62 ± 0.49&39.15 ± 1.31 & 33.90 ± 0.35\\ 
\midrule
CiteSeer &{\bf 51.98} ± 0.03&41.63 ± 0.37&51.83 ± 0.24 & 41.58 ± 0.01\\
\midrule
PubMed &{\bf 29.31} ± 0.12&28.07 ± 0.12&28.66 ± 0.57 & 27.95 ± 0.08\\
\midrule
Collab &{\bf 25.24} ± 0.01&24.76 ± 0.03&24.78 ± 0.69 & 24.80 ± 0.69\\
\bottomrule
\end{tabular}
\label{table:flex_ablation}
\end{adjustbox}
\end{wraptable}

\subsection{RQ2: Framework Ablation} \label{ssec:ablation}

In order to determine which components of FLEX function to improve performance, we ablate across singular mechanisms which are directly involved with the FLEX-tuning process for the co-trained GNN. This includes the use of (a) semi-implicit variation, (b) an expressive link encoder (SEAL), (c) the LP loss $\mathcal{L}_{\text{LP}}$ described in Eq.~\ref{equation:7}. As shown in Table~\ref{table:flex_ablation}, ablating each component leads to a consistent decrease on four different datasets, thus validating the importance of each component.

\subsection{RQ3: Hyperparameter Sensitivity} \label{ssec:hyper}

\begin{figure}[htbp]
    \centering
    \includegraphics[width=0.9\textwidth]{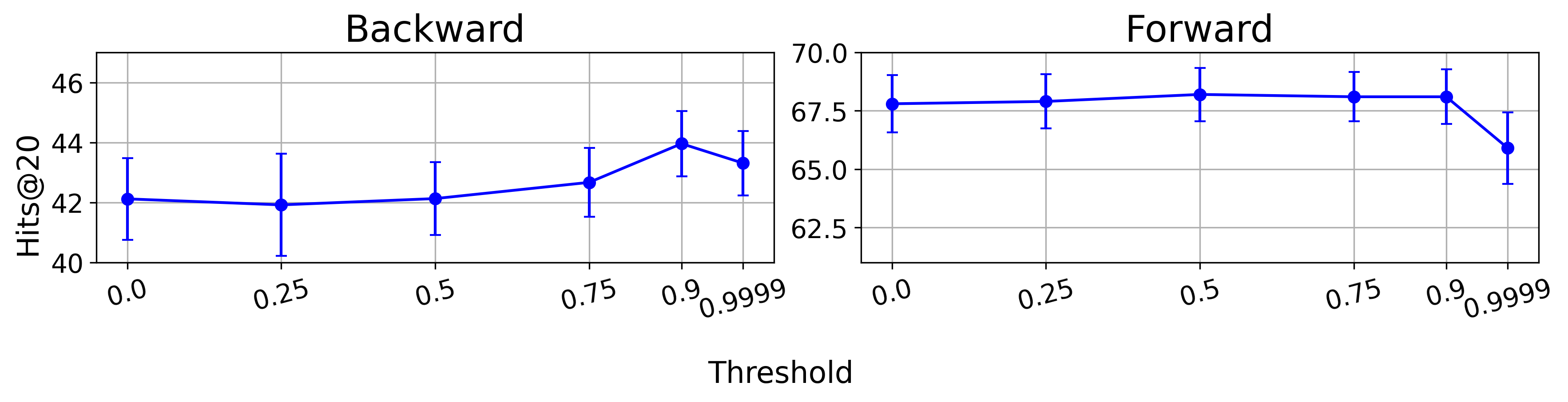}
    \caption{Performance of FLEX on the "Backwards CN" CiteSeer dataset across thresholds. }
    \label{fig:hyper_threshold}
\end{figure}

In order to gauge the impact the that Eq.~\eqref{equation:6} has on downstream performance for FLEX, we conduct a study which measures the difference in performance across the indicator function's target $\gamma = \{0.0, 0.25, 0.5, 0.75, 0.9, 0.9999\} $. As shown in Figure~\ref{fig:hyper_threshold}, we see that the `Backward' split experiences gradually increasing performance up to a value of $0.9$ while the `Forward' split performance sharply decreases at a threshold value of $0.9999$. Given that indicator threshold values directly affect edge-probabilities, these results demonstrate that sparser generated graphs are useful for the 'Backward' split to a point. Whereas little seems to affect a change in the 'Forward' split performance until the graph grows too sparse at $0.9999$. We also include the effect of the learning rate in Figure~\ref{fig:hyp_sens2}.

\subsection{RQ4: OOD Structural Alignment} \label{ssec:struct}

\begin{figure}[htbp]
    \centering
    \includegraphics[width=\textwidth]{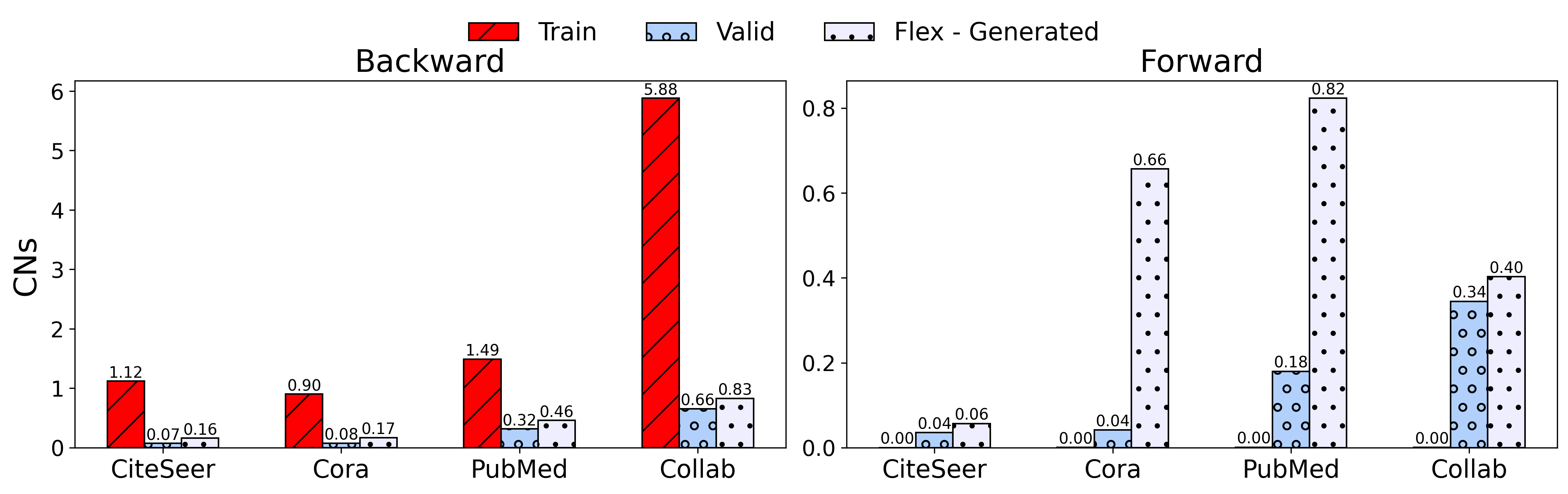}
    \caption{The distribution of Common Neighbors (CNs) scores across different dataset splits for the Backward and Forward CN LPshift splits. }
    \label{fig:heur_full}
\end{figure}

To further verify the effect that FLEX has on graph structure, we directly measure the distribution of CN samples for the original dataset's training and validation samples in comparison to the samples generated by the GGM after tuning within the FLEX framework. We want to verify that FLEX learns to generate useful counterfactuals relevant for link prediction. As shown in Figure~\ref{fig:heur_full}, the 'Flex - Generated' sample distribution closely matches the distribution of validation samples for the 'Backward' subplot, with none of the FLEX samples exceeding a difference of 0.17 CNs. This is a 3-10x improved alignment versus the original training distribution. Within the 'Forward' split, FLEX samples are verifiably denser than the 0 CNs present in training. Despite this, the threshold function still manages to ensure that FLEX samples never exceed a CN threshold of 1. This indicates that FLEX is {\bf successfully targeting structure to produce graphs which are meaningfully different from the training distribution} and help improve performance. A core consideration is FLEX's ability to do this without access to validation or testing samples.

%% file: sections/conclusion.tex

\section{Conclusion}

Within this work, we introduce FLEX, a simple generative framework which targets structure within input graph samples to produce output samples which are verifiably different from training data and more similar to the validation distribution in OOD scenarios. Further experimentation indicates FLEX's ability to model OOD structures without access to validation and testing distributions. Additionally, tuning within the FLEX framework improves performance under realistic and synthetic distribution shifts where traditional generalization methods often decrease performance. This success indicates a pathway for future work for learning to augment graph structure in order to improve performance of LP models in OOD scenarios.

%% file: sections/appendix.tex
\section{HyperParameter Settings} \label{sec:hyper}

Initial tuning of GCN on all tested datasets and NCN on the LPShift datasets followed a hierarchical approach. Initially, GCN was tuned for 1000 epochs in single runs with early-stopping when validation performance did not improve after 20 steps, a learning rate of $1e-3$ and dropout of $0$ across a number of layers $= \{2, 3\}$ and number of hidden channels $= \{128, 256\}$ and batch sizes $= \{32, 64, 128, 256, 512, 1024, 2048, 4096, 8192, 16384, 32768, 65536\}$. Initial NCN tuning followed the same approach, except for being limited to 100 epochs. Dropout and Learning Rate were fixed across the backbone GCN and link predictor.

The second phase of GCN and NCN tuning fixed hidden channels, number of layers, and batch size and then search across a space of learning rate \{1e-2, 1e-3, 1e-4, 1e-5, 1e-6, 1e-7\} and dropout $= \{0.1, 0.3, 0.5, 0.7\}$. NCN was tuned on the ogbl-collab dataset following the author's provided hyperparameters~\cite{wang2023neural}, as indicated in Table~\ref{table:ncn_params} Tuning of the OOD baselines follows the methodology set in~\cite{gui2022good}. Where the tuned GCN has the OOD method applied post-hoc and tuned across their loss coefficients as follows: CORAL $ = \{0.01, 1.0, 0.1\}$, VREx $= \{10.0, 1000.0, 100.0\}$, IRM $= \{10.0, 0.1, 1.0 \}$, DANN $ = \{0.1, 1.0, 0.01\}$, GroupDRO $= \{0.01, 1.0, 0.1\}$. The number of sampled environmental subsets was fixed at $3$ and sampled randomly at program start.

All models, irrespective of FLEX, were evaluted on the full adjacency matrix to ensure consistency with original results.

SIG-VAE, VGAE, and GAE were tuned for 2000 epochs with early stopping set to 100 epochs across learning rates \{1e-3, 1e-4\}. Models were chosen based on their loss values. All generative auto-encoders were fixed to 32 hidden dimensions and 16 output dimensions to model $\mu$, with variation encoders also modeling $\sigma$. The zero-one labeling trick was applied solely to the generative auto-encoder, with a latent embedding size of $(1000, \text{Num. Hidden})$. Given significant time complexity of pre-training SIG-VAE, a random seed was chosen for SIG-VAE and it's respective GNN and then tested across ten unique seeded runs to obtain final performance.

FLEX was tuned for single seeded runs across learning rates $= \{1e-2, 1e-3, 1e-4, 1e-5, 1e-6, 1e-7\}$, threshold values $= \{0.0, 0.25, 0.5, 0.75, 0.9, 0.99, 0.999, 0.9999\}$, and alpha $=\{0.5, 0.7, 0.95, 1.05\}$.

\begin{table}[h]
\centering
\caption{NCN Hyperparameters for the ogbl-collab dataset.}
\begin{adjustbox}{width=0.6\textwidth}
\begin{tabular}{c|c|c|c}

\toprule
\textbf{Parameter} & \textbf{Value} & \textbf{Parameter} & \textbf{Value} \\ 
\midrule
GNN Learning Rate & 0.0082 & Predictor&0.0037\\ 
\midrule
X Dropout& 0.25& T Dropout & 0.05\\
\midrule
PT & 0.1& GNN EdgeDropout & 0.25\\
\midrule
Predictor Edge Dropout & 0.0 & Predictor Dropout & 0.3\\
\midrule
GNN Dropout & 0.1 & Probability Scaling & 2.5\\
\midrule
Probability Offset & 6.0 &Alpha& 1.05\\
\midrule
Batch Size & 65536 & Layer Norm & True \\
\midrule
Layer Norm N & True & Predictor & GCN \\
\midrule
Epochs & 100 & Model & GCN \\
\midrule
Hidden Dimension & 64 & MP Layers & 1\\
\midrule
Test Batch Size & 131072 & Mask Input & True \\
\midrule
Validation Edges As Input & True & Res. & True \\
\midrule
Use X. Linear & True & Tail Acting & True \\
\bottomrule
\end{tabular}
\label{table:ncn_params}
\end{adjustbox}
\end{table}

\section{Synthetic Dataset Split Settings} \label{sec:synth}

LPShift datasets were generated following the process described by the authors in~\cite{revolinsky2024understanding}. They consider three types of datasets splits that divide the links based on common heuristics. This includes: {\it CN} = Common Neighbors~\cite{adamic2003friends}, {\it SP} = Shortest-Path, {\it PA} = Preferential-Attachment~\cite{liben2003link}. They further include two ``directions'' for how the links are split. A `Forward' splits indicates that the value of the heuristics increase from train to valid and then test. The `Backwards` split indicates that they decrease. The splits are defined based on two threshold parameters. For the `Forward' splits the first parameter defines the upper-bound on training data and the second the lower-bound on testing data. The opposite is true for the `Backwards' split. For example, the CN split of `1, 2' indicates that training links contain CNs in the range $[0, 1)$, valid in $[1, 2)$, and test $[2, \infty)$. For a CN split of `2, 1', the training and testing links would be flipped. The parameters used across all tested LPShift datasets are detailed below in Table~\ref{table:lpshift} and follow those used by the original authors~\cite{revolinsky2024understanding}. Note that these are the same across all datasets used.

\begin{table}[h]
\centering
\caption{LPShift Dataset Parameters.}
\begin{adjustbox}{width=0.75\textwidth}
\begin{tabular}{c|c|c|c}

\toprule
\textbf{'Backward' Split} & \textbf{Parameters} &\textbf{'Forward' Split} & \textbf{Parameters} \\ 
\midrule
SP & 26, 17 & SP & 17, 26\\
\midrule
CN & 2,1 & CN & 1,2\\
\midrule
PA & 50, 100 & PA & 100, 50\\
\bottomrule
\end{tabular}
\label{table:lpshift}
\end{adjustbox}
\end{table}

\section{Resources} \label{sec:resources}

All models and datasets were tuned and tested on Nvidia A5000 GPUs with 24 GB available RAM and a server with 128 cores and 1TB available RAM.

\section{Hyper-Parameter Sensitivity} \label{sec:sensi}

\begin{figure*}[h]
    \centering
    \includegraphics[width=\textwidth]{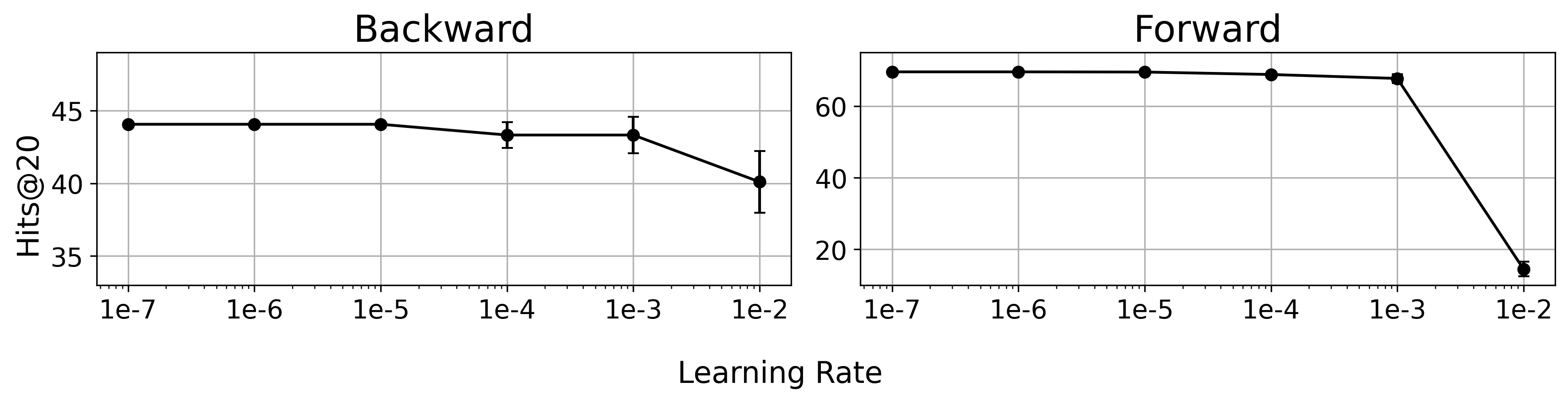}
    \caption{The Hits@20 Scores for FLEX on the "Backwards" - CN CiteSeer Dataset across different learning rates.}
    \label{fig:hyp_sens2}
\end{figure*}

Within this section, we indicate how learning rate affect performance within the FLEX framework. As shown in both the 'Backward' and 'Forward' subplots in Figure~\ref{fig:hyp_sens2} a higher learning rate contributes to monotonically decreasing performance. This represents a potential pitfall when FLEX-tuning any pre-trained GNNs. Especially since FLEX relies on subgraph samples whereas GNNs often train on a full adjacency matrix.

\section{Flex Algorithms} \label{sec:node_dec}

As defined earlier in Section~\ref{sec:flex_model}, FLEX operates in two critical stages, (1): The generative graph model (GGM) is pretrained on labeled subgraphs extracted from the target dataset following Eq.~\eqref{equation:2}. While the GNN is pre-training separately on the full adjacency matrix. This is defined on lines 3-5 in Algorithm~\ref{alg:FLEX}. (2): After pre-training, the generative GGM is then placed within the FLEX framework and co-trained with the GNN following Eq.~\eqref{equation:7}. At each subsequent mini-batch, the GGM produces new synthetic graphs and therefore new structural views of the original dataset which are subsequently passed into the GNN to gauge sample validity. This is defined on lines 6-10 in Algorithm~\ref{alg:FLEX}. Given that the divergence between the posterior and prior distributions is maximized, this means that subsequent epochs should converge to generate a final distribution that is structurally different from the training samples. As mentioned in Section~\ref{sec:node_decode}, Algorithm~\ref{alg:SubgraphVAE} takes in feature input and a representative block-diagonal matrix to ensure that SIG-VAE is expressive to mini-batch samples of varying node numbers~\cite{hasanzadeh2019semi}.

\begin{algorithm}[H] 
\caption{FLEX - Pre-training and Tuning} \label{alg:FLEX}
\begin{algorithmic}[1] 
\Require $\mathbf{G(X, A)}$, $\mathbf{X} \in\mathbb{R}^{N\times d}$
\State Extract $\mathbf{G_s}$ from 1-hop enclosed subgraphs of $A$
\State Retrieve $\mathbf{Z}$ using the zero-one labeling trick, Eq.~\eqref{equation:5}
\For{epoch = 1 to pretrain}
    \State Train SIG-VAE on $\mathbf{G_s}$ using Eq.~\eqref{equation:2} and labels $\mathbf{Z}$
\EndFor
\For{epoch = 1 to flex-tune}
    \State Sample $\mathbf{G_s'}$ from SIG-VAE
    \State Apply Eq.~\eqref{equation:5} on $\mathbf{G_s'}$
    \State Train GNN + SIG-VAE on $\mathbf{G_s'}$ using Eq.~\eqref{equation:7}
\EndFor
\end{algorithmic}
\end{algorithm}

\begin{algorithm}[H] 
\caption{Node-Aware Decoder Algorithm} \label{alg:SubgraphVAE}
\begin{algorithmic}[1]

\Require
    \Statex $x \in \mathbb{R}^{N \times F}$: Node features, $A = \mathrm{diag}(A_1, \dots, A_K)$: Block-diagonal adjacency
    \Statex $\mathbf{Z} \in \mathbb{R}^{N \times d}$: Structural features, $J$: Truncation index, $n_{train} = [\mathcal{N}_1, \dots, \mathcal{N}_N]$: Training Nodes
\newline
\State $(\mu, \log \sigma^2, \mathrm{SNR}) \gets \mathrm{Encoder}(x, A, Z)$
\State $\mu' \gets \mu_{J:N}$,\quad $\log \sigma'^2 \gets \log \sigma^2_{J:N}$

\State Split $\mu', \log \sigma'^2$ into subgraphs $\mu_i, \log \sigma_i^2$ using $n_{train}$
\For{$i = 1$ to $N$}
    \State Sample $\epsilon_i \sim \mathcal{N}(0, I)$
    \State $z_i \gets \mu_i + \epsilon_i \odot \exp(0.5 \cdot \log \sigma_i^2)$ \Comment{Reparametrization Trick}
    \State $(\hat{A}_i, z_i^{\text{scaled}}, r_k) \gets \mathrm{Decoder}(z_i)$ 
    \State Insert $\hat{A}_i$ into $\hat{A}_{\text{global}}$ at block $(i,i)$
    \State Insert $z_i$, $z_i^{\text{scaled}}$, $\epsilon_i$ into global tensors
\EndFor

\State \Return $\hat{A}_{\text{global}}, \mu, \log \sigma^2, \mathbf{Z}_{\text{global}}, \mathbf{Z}_{\text{global}}^{\text{scaled}}, \epsilon_{\text{global}}, r_k, \mathrm{SNR}$
\end{algorithmic}
\label{alg:forward_pass}
\end{algorithm}

\section{Degree Bias Investigation} \label{sec:app_deg}

\begin{figure*}[h]
    \centering
    \includegraphics[width=0.5\textwidth]{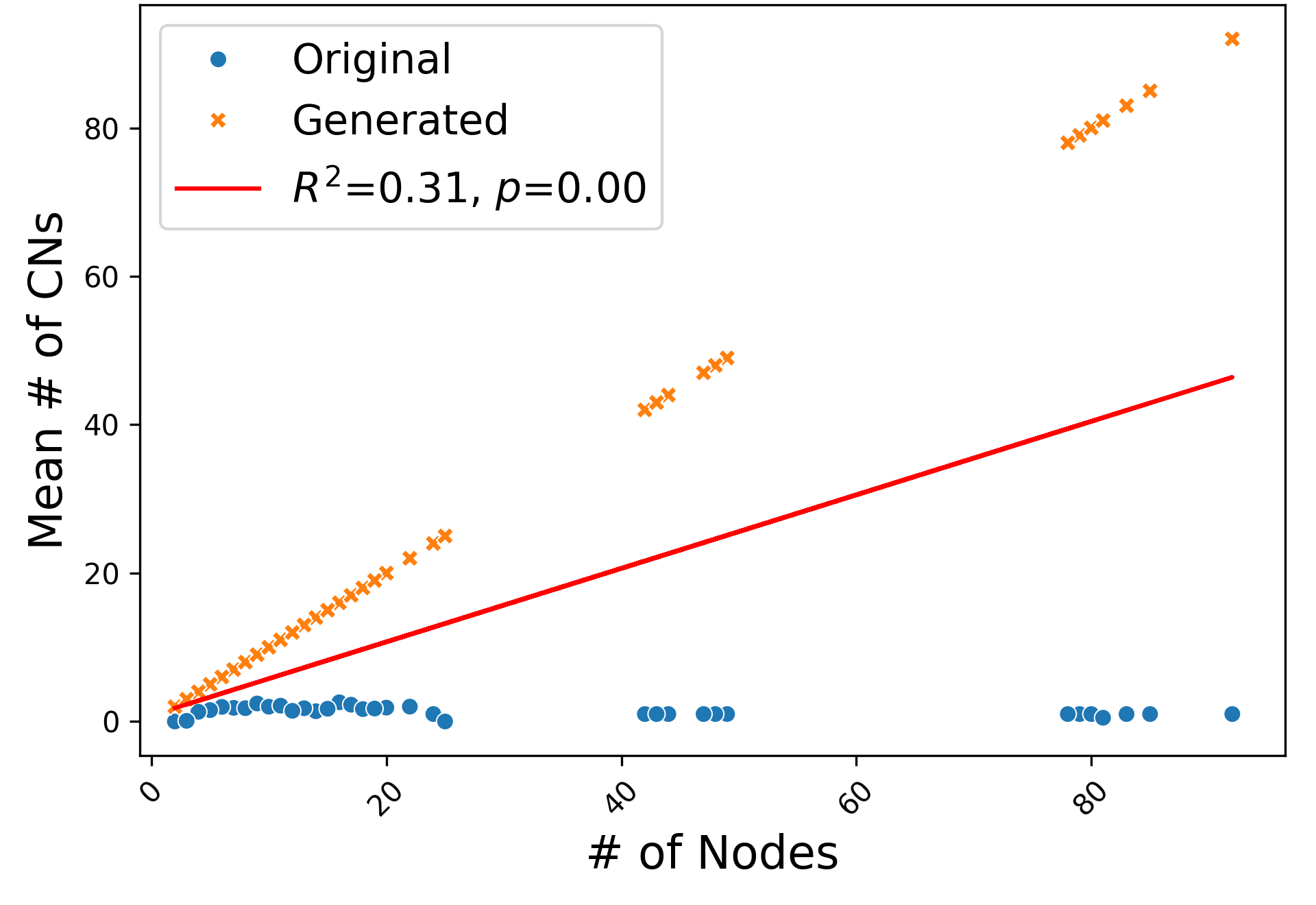}
    \caption{The distribution of Mean Common Neighbors and Mean Number of Nodes for subgraph samples generated by FLEX on the 'Backward' LPShift CN - Cora dataset without the threshold function. Note the near-perfect linear growth of Common Neighbors with respect to the number of nodes within a given input subgraph.}
    \label{fig:0ThreshCora}
\end{figure*}

\begin{figure*}[h]
    \centering
    \includegraphics[width=0.5\textwidth]{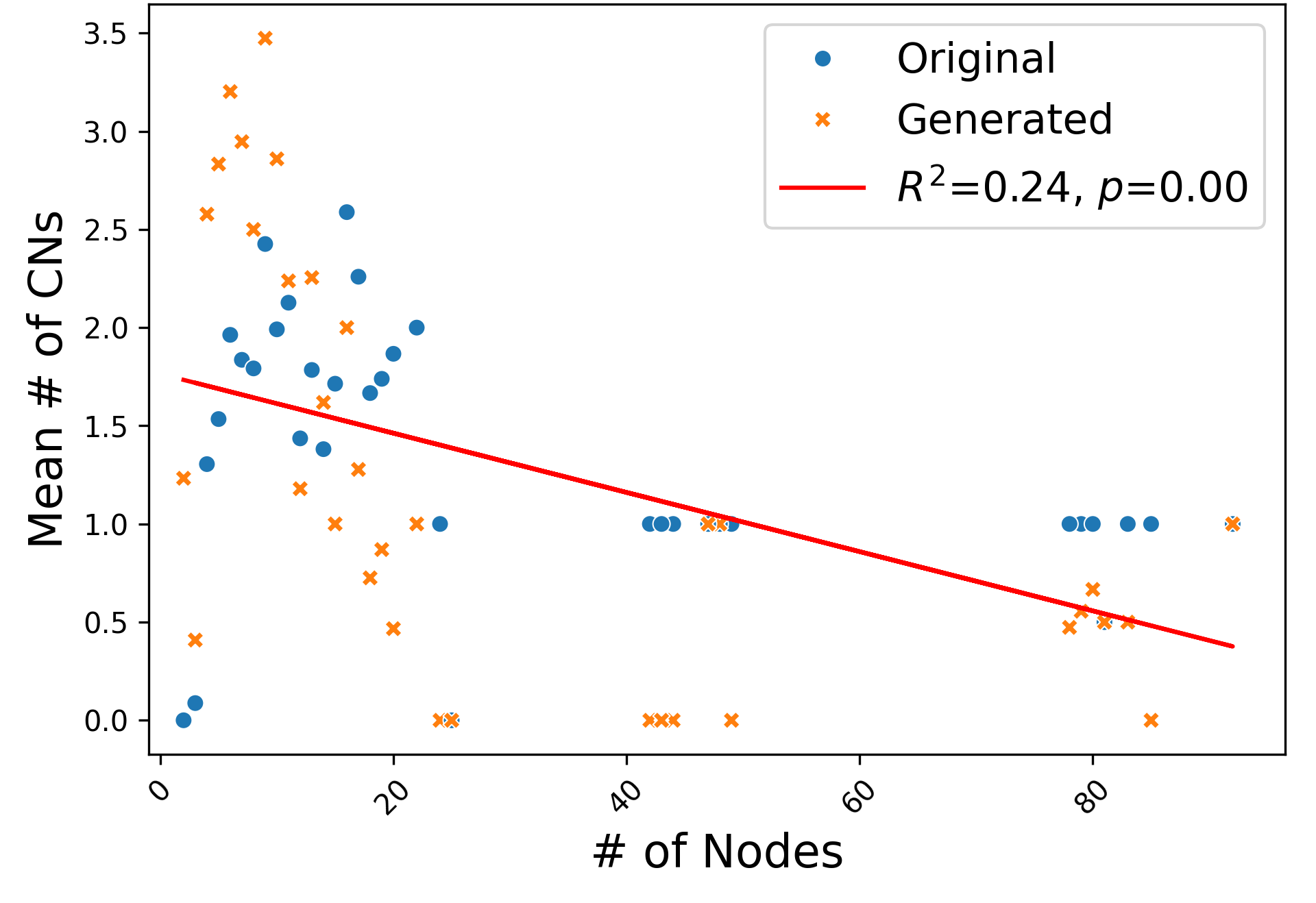}
    \caption{The distribution of Mean Common Neighbors and Mean Number of Nodes for subgraph samples generated by FLEX on the 'Backward' LPShift CN - Cora dataset after applying the threshold function. The threshold function ensures that low-probabilities edges are not formed, resulting in generated samples with a common neighbors that are morely closely-correlated to the input samples.}
    \label{fig:9999ThreshCora}
\end{figure*}

As previously-mentioned in Section~\ref{sec:node_decode}, the generated subgraph samples without an indicated threshold suffer from degree-bias~\cite{tang2020investigating}, thereby resulting in densely-generated outputs even on sparse inputs. This effect is demonstrated in Figure~\ref{fig:0ThreshCora}, as shown with the perfect linear relationship between the mean number of common neighbors in the output sample respective to the number of nodes within input samples. To combat this, the indicator threshold is tuned to eliminate edge-probabilities with a lower threshold than indicated. The effect of this threshold can be seen in Figure~\ref{fig:9999ThreshCora}, where a threshold of $0.9999$ reduces the maximum mean number of Common Neighbors by a factor of 40, as respective to Figure~\ref{fig:0ThreshCora}. This then shows a more meaningful correlation between output CNs and input nodes, meaning that output graphs are no longer densely-connected which serves as a desirable property when attempt to generalize on much sparser graphs; like those contained within the 'Backward' CN Cora dataset.

\clearpage

\section{Dataset Licenses} \label{sec:licenses}

Both OGB~\cite{hu2020open} and LPShift~\cite{revolinsky2024understanding}, the datasets considered in our study, are licensed under the MIT license.

\section{Limitations} \label{sec:limitations}

From a theoretical perspective, FLEX operates under the critical assumption that there are counterfactual substructures which exist under the causal model that constructed the original dataset. If no such substructures are present, (i.e. the dataset samples are not OOD), then FLEX is also likely to decrease model performance.

For practical implementation, FLEX requires sampling $k$-hop enclosed subgraphs, which can be computationally-restrictive if applied with settings that are normal when training on full adjacency matrices. Additionally, if poorly-tuned then SIG-VAE will produce meaningless outputs and decrease downstream performance regardless of how well pre-trained the GNN is. FLEX is also subject to over-tuning, where a single epoch can increase performance but subsequent epochs can lead to a monotonic decrease in performance. 

\section{Societal Impact} \label{sec:impact}

Our proposed method, FLEX, aims to improve the generalization capabilities of link prediction methods. Since generalization is a key real-world concerns for many ML models, we argue that FLEX has a potential to have a positive impact. Furthermore, link prediction is a common task used in many fields such as recommender systems, drug-drug interactions, and knowledge graph reasoning. Thus, improving the generalization of link prediction in those fields can be helpful for future research. Therefore, no apparent risk is related to the contribution of this work.